\documentclass[english]{llncs}
\usepackage{babel}
\usepackage{array}
\usepackage{multirow}
\usepackage{amsmath}
\usepackage{graphicx}
\usepackage[unicode=true,pdfusetitle,
 bookmarks=true,bookmarksnumbered=false,bookmarksopen=false,
 breaklinks=false,pdfborder={0 0 1},backref=false,colorlinks=false]
 {hyperref}

\makeatletter

\providecommand{\tabularnewline}{\\}

\usepackage{microtype}

\makeatother

\begin{document}

\title{Outlier Detection on Mixed-Type Data: An Energy-based Approach}

\author{Kien Do, Truyen Tran, Dinh Phung and Svetha Venkatesh}

\institute{Centre for Pattern Recognition and Data Analytics\\
Deakin University, Geelong, Australia\\
\emph{dkdo@deakin.edu.au}}

\maketitle
\global\long\def\xb{\boldsymbol{x}}
\global\long\def\yb{\boldsymbol{y}}
\global\long\def\eb{\boldsymbol{e}}
\global\long\def\zb{\boldsymbol{z}}
\global\long\def\hb{\boldsymbol{h}}
\global\long\def\ab{\boldsymbol{a}}
\global\long\def\bb{\boldsymbol{b}}
\global\long\def\cb{\boldsymbol{c}}
\global\long\def\sigmab{\boldsymbol{\sigma}}
\global\long\def\gammab{\boldsymbol{\gamma}}
\global\long\def\alphab{\boldsymbol{\alpha}}
\global\long\def\rb{\boldsymbol{r}}
\global\long\def\fb{\boldsymbol{f}}
\global\long\def\ib{\boldsymbol{i}}
\global\long\def\Wb{\boldsymbol{W}}
\global\long\def\pb{\boldsymbol{p}}

\begin{abstract}
Outlier detection amounts to finding data points that differ significantly
from the norm. Classic outlier detection methods are largely designed
for single data type such as continuous or discrete. However, real
world data is increasingly heterogeneous, where a data point can have
both discrete and continuous attributes. Handling mixed-type data
in a disciplined way remains a great challenge. In this paper, we
propose a new unsupervised outlier detection method for mixed-type
data based on Mixed-variate Restricted Boltzmann Machine (Mv.RBM).
The Mv.RBM is a principled probabilistic method that models data density.
We propose to use \emph{free-energy} derived from Mv.RBM as outlier
score to detect outliers as those data points lying in low density
regions. The method is fast to learn and compute, is scalable to massive
datasets. At the same time, the outlier score is identical to data
negative log-density up-to an additive constant. We evaluate the proposed
method on synthetic and real-world datasets and demonstrate that (a)
a proper handling mixed-types is necessary in outlier detection, and
(b) free-energy of Mv.RBM is a powerful and efficient outlier scoring
method, which is highly competitive against state-of-the-arts.

\end{abstract}

\section{Introduction}

Outliers are those deviating significantly from the norm. Outlier
detection has broad applications in many fields such as security \cite{diehl2002real,kruegel2003anomaly,portnoy2001intrusion},
healthcare \cite{tran2013integrated}, and insurance \cite{konijn2011finding}.
A common assumption is that outliers lie in the low density regions
\cite{chandola2009anomaly}. Methods implementing this assumption
differ in how the notion of density is defined. For example, in nearest
neighbor methods ($k$-NN) \cite{angiulli2002fast}, large distance
between a point to its nearest neighbors indicates isolation, and
hence, it lies in a low density region. Gaussian mixture models (GMM),
on the other hand, estimate the density directly through a parametric
family of clusters \cite{mclachlan1988mixture}.

A real-world challenge rarely addressed in outlier detection is mixed-type
data, where each data attribute can be any type such as continuous,
binary, count or nominal. Most existing methods, however, assume homogeneous
data types. Gaussian mixture models, for instance, require data to
be continuous and normally distributed. One approach to mixed-type
data is to reuse existing methods. For example, we can transform multiple
types into a single type \textendash{} the process known as coding
in the literature. A typical practice of coding nominal data is to
use a set of binary variables with exactly one active element. But
it leads to information loss because the derived binary variables
are considered independent in subsequent analysis. Another drawback
of coding is that numerical methods such as GMM and PCA ignore the
binary nature of the derived variables. Another way of model reusing
is to modify existing methods to accommodate multiple types. However,
the modification is often heuristic. Distance-based methods would
define type-specific distances, then combine these distances into
a single measure. Because type-specific distances differ in scale
and semantics, finding a suitable combination is non-trivial. 

A disciplined approach to mixed-type outlier detection demands three
criteria to be met: (i) c\emph{apturing correlation structure between
types}, (ii) \emph{measuring deviation from the norm}, and (iii) \emph{efficient
to compute} \cite{lud2016iscovering}. To this end, we propose a new
approach that models multiple types directly and at the same time,
provides a fast mechanism for identifying low density regions. To
be more precise, we adapt and extend a recent method called Mixed-variate
Restricted Boltzmann Machine (Mv.RBM) \cite{Truyen:2011b}. Mv.RBM
is a generalization of the classic RBM \textendash{} originally designed
for binary data, and now a building block for many deep learning architectures
\cite{bengio2013representation,hinton2006rdd}. Mv.RBM has been applied
for representing \emph{regularities} in survey analysis \cite{Truyen:2011b},
multimedia \cite{tu_truyen_phung_venkatesh_icme13} and healthcare
\cite{tu_truyen_phung_venkatesh_pakdd13}, but not for outlier detection,
which searches for \emph{irregularities}. Mv.RBM captures the correlation
structure between types through factoring \textendash{} data types
are assumed to be independent given a generating mechanism. 

In this work, we extend the Mv.RBM to cover counts, which are then
modeled as Poisson distribution \cite{salakhutdinov2009semantic}.
We then propose to use \emph{free-energy }as outlier score to rank
mixed-type instances. Note that \emph{free-energy} is notion rarely
seen in outlier detection. In RBM, free-energy equals the negative
log-density up to an additive constant, and thus offering a principled
way for density-based outlier detection. Importantly, estimation of
Mv.RBM is very efficient, and scalable to massive datasets. Likewise,
free-energy is computed easily through a single matrix projection.
Thus Mv.RBM coupled with free-energy meets all the three criteria
outlined above for outlier detection. We validate the proposed approach
through an extensive set of synthetic and real experiments against
well-known baselines, which include the classic single-type methods
(PCA, GMM and one-class SVM), as well as state-of-the-art mixed-type
methods (ODMAD \cite{koufakou2008detecting}, Beta mixture model (BMM)
\cite{bouguessa2015practical} and GLM-t \cite{lu2016discovering}).
The experiments demonstrate that (a) a proper handling mixed-types
is necessary in outlier detection, and (b) free-energy of Mv.RBM is
a powerful and efficient outlier scoring method, being highly competitive
against state-of-the-arts.

In summary, we claim the following contributions:
\begin{itemize}
\item Introduction of a new outlier detection method for mixed-type data.
The method is based on the concept of free-energy derived from a recent
method known as Mixed-variate Restricted Boltzmann Machine (Mv.RBM).
The method is theoretically motivated and efficient.
\item Extension of Mv.RBM to handle counts as Poisson distribution.
\item A comprehensive evaluation on synthetic and real mixed-type datasets,
demonstrating the effectiveness of the proposed method against classic
and state-of-the-art rivals.
\end{itemize}

\section{Related Work}

Outliers, also known as anomalies or novelties, are those thought
to be generated from a mechanism different from the majority. Outlier
detection is to recognize data points with unusual characteristics,
or in other word, instances that do not follow any regular patterns.
When there is very little or no information about outliers provided,
which is common in real world data, the regular patterns need to be
discovered from normal data itself. This is called \emph{unsupervised}
anomaly detection. A variant known as \emph{semi-supervised} is when
the training data is composed of just normal data \cite{chandola2009anomaly}.

\subsubsection{Single Type Outlier Detection}

A wide range of unsupervised methods have been proposed, for example,
distance-based (e.g., $k$-NN \cite{angiulli2002fast}), density-based
(e.g., LOF \cite{breunig2000lof}, LOCI \cite{papadimitriou2003loci}),
cluster-based (e.g., Gaussian mixture model or GMM), projection-based
(e.g., PCA) and max margin (One-class SVM). Distance-based and density-based
methods model the local behaviors around each data point at a high
level of granularity while cluster-based methods group similar data
points together into clusters \cite{aggarwal2015outlier}. Projection-based
methods, on the other hand, find a data projection that is sensitive
to outliers. A comprehensive review of these methods were conducted
by Chandola \textit{et al}. \cite{chandola2009anomaly}.

\subsubsection{Mixed-Type Outlier Detection}

Although pervasive in real-world domains, mixed-type data is rarely
addressed in the literature. When data is mixed (e.g., continuous
and discrete), measuring distance between two data points or estimating
data density can be highly challenging. A naïve solution is to transform
mixed-types into a single type, e.g., by coding nominal variables
into 0/1 or discretizing continuous variables. This practice can significantly
distort the true underlying data distribution and result in poor performance
\cite{koufakou2008detecting}. In order to handle mixed-type data
directly, several methods have been proposed. LOADED \cite{ghoting2004loaded}
uses frequent pattern mining to define the score of each data point
in the nominal attribute space and link it with a precomputed correlation
matrix for each item set in continuous attribute space. Since there
are a large number of item sets generated, this method suffers from
high memory cost. RELOAD \cite{otey2005fast} is a memory-efficient
version of LOADED, which employs a set of Naïve Bayes classifiers
with continuous attributes as inputs to predict abnormality of nominal
attributes instead of aggregating over a large number of item sets.

Koufakou \textit{et al}. \cite{koufakou2008detecting} propose a method
named ODMAD to detect outliers in sparse data with both nominal and
continuous attributes. Their method first computes the anomaly score
for nominal attributes using the same algorithm as LOADED. Points
detected as outliers at this step are set aside and the remaining
are examined over continuous attribute space with cosine similarity
as a measurement. In \cite{bouguessa2015practical}, separate scores
over nominal data space and numerical data space are calculated for
each data point. The list of two dimensional score vectors of data
was then modeled by a mixture of bivariate beta distributions. Similar
to other cluster-based methods, abnormal objects could be detected
as having a small probability of belonging to any components. Although
the idea of beta modeling is interesting, the calculation of scores
is still very simple, which is $k$-NN distance for continuous attributes
and sum of item frequencies for nominal attributes.

The work of \cite{zhang2010effective} adopts a different approach
called Pattern-based Outlier Detection (POD). A pattern is a subspace
formed by a particular nominal fields and all continuous fields. A
logistic classifier is trained for each subspace pattern, in which
continuous and nominal attributes are explanatory and response variables,
respectively. The probability returned by the classifier measures
the degree to which an instance deviates from a specific pattern.
This is called Categorical Outlier Factor (COF). The collection of
COFs and $k$-NN distance form the final anomaly score for a data
example. Given a nominal attribute, POD models the functional relationship
between continuous variables. The dependency between nominal attributes,
however, is not actually captured. Moreover, when data only contains
nominal attributes, the classifier cannot be created. 

For all the methods mentioned above, their common drawback is that
they are only able to capture correlation between a set of nominal
and numerical attributes but not pair-wise correlations. The most
recent work of Lu \textit{et al.} \cite{lu2016discovering} overcomes
the mentioned drawback and models the data distribution. \textit{\emph{They
design a Generalized Linear Model framework accompanied with a latent
variable for correlation capturing and an another latent variable
following Student-t distribution as an error buffer. The main advantage
of this method is that it provides strong a statistical foundation
for modeling distribution of different types. However, the inference
for detecting outliers is inexact and expensive to compute.}}

\subsubsection{Restricted Boltzmann Machine}

(RBM) is a probabilistic model of binary data, formulated as a bipartite
Markov random field. This special structure allows efficient inference
and learning \cite{Hinton02}. More recently, it was used as a building
block for Deep Belief Networks \cite{hinton2006rdd}, the work that
started the current revolution of deep learning \cite{lecun2015deep}.
Recently RBM has been used for single-type outlier detection \cite{fiore2013network}.

\section{Mixed-Type Outlier Detection}

In this section, we present a new density-based method for mixed-type
outlier detection. Given a data instance $\xb$ we estimate the density
$P(\xb)$ then detect if the instance is an outlier using a threshold
on the density:
\begin{equation}
-\log P(\xb)\ge\beta\label{eq:outlier-decision}
\end{equation}
for some predefined threshold $\beta$. Here $-\log P(\xb)$ serves
as the outlier scoring function.

\subsection{Density Estimation for Mixed Data}

Estimating $P(\xb)$ is non-trivial in mixed-type data since we need
to model correlation structures within-type and between-types. A direct
correlation between-types demands a careful specification for each
type-pair. For example, for two variables of different types $x_{1}$
and $x_{2}$, we need to specify either $P(x_{1},x_{2})=P(x_{1})P(x_{2}\mid x_{1})$
or $P(x_{1},x_{2})=P(x_{2})P(x_{1}\mid x_{2})$. With this strategy,
the number of pairs grows quadratically with the number of types.
Most existing methods follow this approach and they are designed for
a specific pair such as binary and Gaussian \cite{de2013analysis}.
They neither scale to large-scale problems nor support arbitrary types
such as binary, continuous, nominal, and count. 

Mixed-variate Restricted Boltzmann Machine (Mv.RBM) is a recent method
that supports arbitrary types simultaneously \cite{Truyen:2011b}.
It bypasses the problems with detailed specifications and quadratic
complexity by using latent binary variables. Correlation between types
is not modeled directly but is factored into indirect correlation
with latent variables. As such we need only to model the correlation
between a type and the latent binary. This scales linearly with the
number of types. 

Mv.RBM was primarily designed for data representation which transforms
mixed data into a homogeneous representation, which serves as input
for the next analysis stage. Our adaptation, on the other hand, proposes
to use Mv.RBM as outlier detector directly, without going through
the representation stage.

\begin{figure}
\begin{centering}
\includegraphics[width=0.6\textwidth]{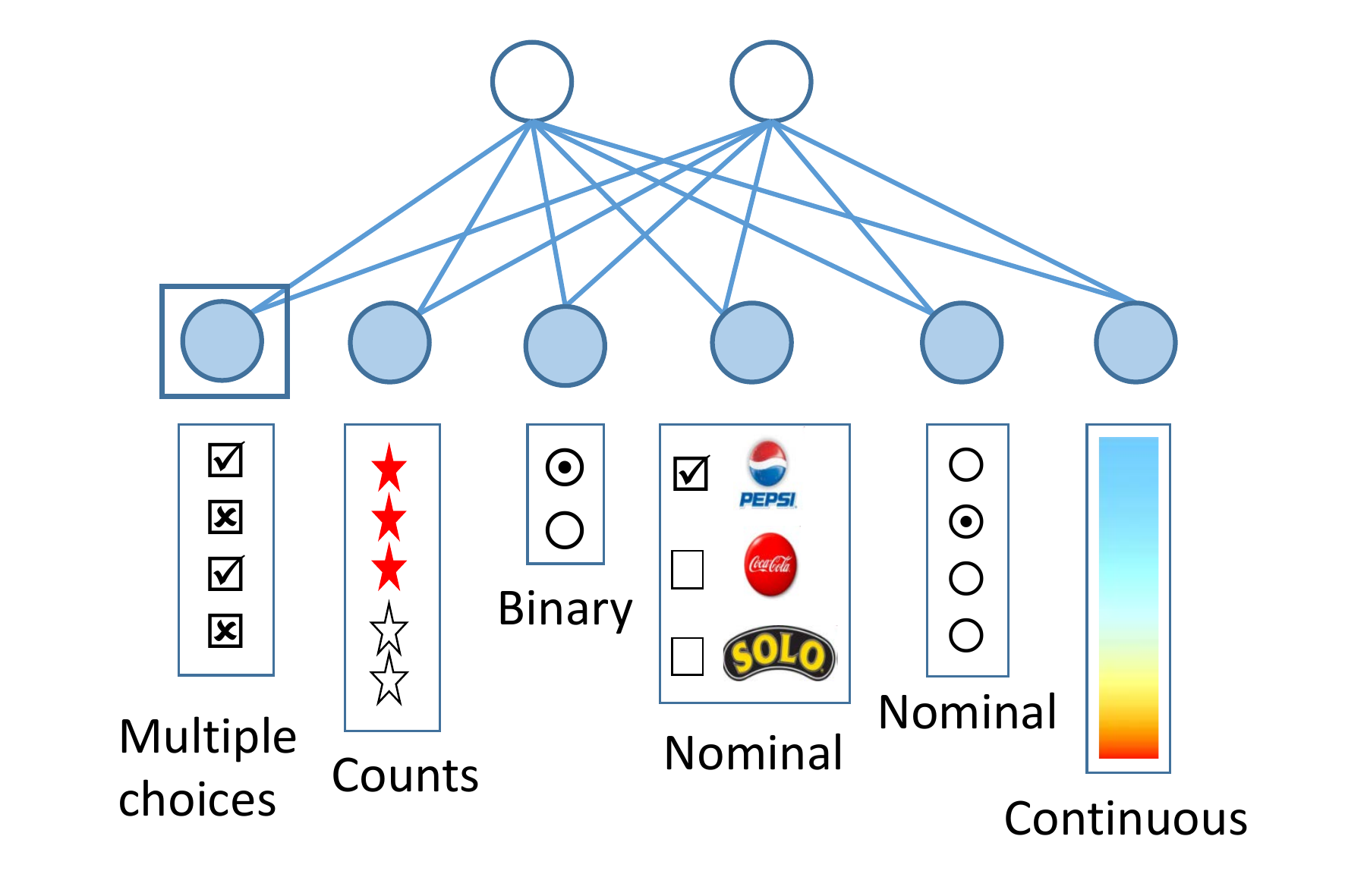}
\par\end{centering}
\caption{Mix-variate Restricted Boltzmann machines for mixed-type data. Filled
circles denote visible inputs, empty circles denote hidden units.
Multiple choices are modeled as multiple binaries, denoted by a filled
circle in a clear box.\label{fig:model}}
\end{figure}

\subsection{Mixed-variate Restricted Boltzmann Machines}

We first review Mv.RBM for a mixture of binary, Gaussian and nominal
types, then extend to cover counts. See Fig.~\ref{fig:model} for
a graphical illustration. Mv.RBM is an extension of RBM for multiple
data types. An RBM is a probabilistic neural network that models binary
data in an unsupervised manner. More formally, let $\xb\in\{0,1\}^{N}$
be a binary input vector, and $\hb\in\{0,1\}^{K}$ be a binary hidden
vector, RBM defines the joint distribution as follows:
\[
P\left(\xb,\hb\right)\propto\exp\left(-E(\xb,\hb)\right)
\]
where $E(\xb,\hb)$ is energy function of the following form:
\begin{equation}
E(\xb,\hb)=-\left(\sum_{i}a_{i}x_{i}+\sum_{k}b_{k}h_{k}+\sum_{ik}W_{ik}x_{i}h_{k}\right)\label{eq:RBM-energy}
\end{equation}
Here $(\ab,\bb,W)$ are model parameters. 

For subsequent development, we rewrite the energy function as:
\begin{equation}
E(\xb,\hb)=\sum_{i}E_{i}(x_{i})+\sum_{k}\left(-b_{k}+\sum_{i}G_{ik}(x_{i})\right)h_{k}\label{eq:Mv.RBM-energy}
\end{equation}
where $E_{i}(x_{i})=-a_{i}x_{i}$ and $G_{ik}(x_{i})=-W_{ik}x_{i}$.

Mv.RBM extends RBM by redefining the energy function to fit multiple
data types. The energy function of Mv.RBM differs from that of RBM
by the using multiple type-specific energy sub-functions $E_{i}(x_{i})$
and $G_{ik}(x_{i})$ as listed\footnote{The original Mv.RBM also covers rank, but we do not consider in this
paper.} in Table~\ref{tab:Type-energy-sub-function}. The energy decomposition
in Eq.~(\ref{eq:Mv.RBM-energy}) remains unchanged.

\begin{table}
\centering{}%
\begin{tabular}{|c|cccc|}
\hline 
Func. & Binary & Gaussian & Nominal & Count\tabularnewline
\hline 
$E_{i}(x_{i})$ & $-a_{i}x_{i}$ & $\frac{x_{i}^{2}}{2}-a_{i}x_{i}$ & $-\sum_{c}a_{ic}\delta(x_{i},c)$ & $\log x_{i}!-a_{i}x_{i}$\tabularnewline
$G_{ik}(x_{i})$ & $-W_{ik}x_{i}$ & $-W_{ik}x_{i}$ & $-\sum_{c}W_{ikc}\delta(x_{i},c)$ & $-W_{ik}x_{i}$\tabularnewline
\hline 
\end{tabular}\caption{Type-specific energy sub-functions. Here $\delta(x_{i},c)$ is the
identity function, that is, $\delta(x_{i},c)=1$ if $x_{i}=c$, and
$\delta(x_{i},c)=0$ otherwise. For Gaussian, we assume data has unit
variance. Multiple choices are modeled as multiple binaries.\label{tab:Type-energy-sub-function}}
\end{table}

\subsubsection{Extending Mv.RBM for Counts}

We employ Poisson distributions for counts \cite{salakhutdinov2009semantic}.
The sub-energy sub-functions are defined as:
\begin{equation}
E_{i}(x_{i})=\log x_{i}!-a_{i}x_{i};\quad\quad G_{ik}(x_{i})=-W_{ik}x_{i}\label{eq:Poisson-type-energy-1}
\end{equation}
Note that count modeling was not introduced in the original Mv.RBM
work.

\subsubsection{Learning}

Model estimation in RBM and Mv.RBM amounts to maximize data likelihood
with respect to model parameters. It is typically done by $n$-step
Contrastive Divergence (CD-$n$), which is an approximate but fast
method. In particular, for each parameter update, CD-$n$ maintains
a very short Mote Carlo Markov chain (MCMC), starting from the data,
runs for $n$ steps, then collects the samples to approximate data
statistics. The MCMC is efficient because of the factorizations in
Eq.~(\ref{eq:factorization}), that is, we can sample all hidden
variables in parallel through $\hat{\hb}\sim P\left(\hb\mid\xb\right)$
and all visible variables in parallel through $\hat{\xb}\sim P\left(\xb\mid\hb\right)$.
For example, for Gaussian inputs, the parameters are updated as follows:
\begin{align*}
b_{k} & \leftarrow b_{k}+\eta\left(\bar{h}_{k\mid\xb}-\bar{h}_{k\mid\hat{\xb}}\right)\\
a_{i} & \leftarrow a_{i}+\eta\left(x_{i}-\hat{x_{i}}\right)\\
W_{ik} & \leftarrow W_{ik}+\eta\left(x_{i}\bar{h}_{k\mid\xb}-\hat{x}_{i}\bar{h}_{k\mid\hat{\xb}}\right)
\end{align*}
where $\bar{h}_{k\mid\xb}=P\left(h_{k}=1\mid\xb\right)$ and $\eta>0$
is the learning rate. This learning procedure scales linearly with
$n$ and data size. 

\subsubsection{Mv.RBM as a Mixture Model of Exponential Size}

In Mv.RBM, types are not correlated directly but through the common
hidden layer. The posterior $P\left(\hb\mid\xb\right)$ and data generative
process $P\left(\xb\mid\hb\right)$ in Mv.RBM are factorized as:
\begin{equation}
P\left(\hb\mid\xb\right)=\prod_{k}P\left(h_{k}\mid\xb\right);\quad\quad P\left(\xb\mid\hb\right)=\prod_{i}P\left(x_{i}\mid\hb\right)\label{eq:factorization}
\end{equation}
Here types are conditionally independent given $\hb$, but since $\hb$
are hidden, types are dependent as in $P(\xb)=\sum_{\hb}P(\xb,\hb)$. 

The posterior has the same form across types \textendash{} the activation
probability $P\left(h_{k}=1\mid\xb\right)$ is $\text{sigmoid}\left(b_{k}-\sum_{i}G_{ik}(x_{i})\right)$.
On the other hand, the generative process is type-specific. For example,
for binary data, the activation probability $P\left(x_{i}=1\mid\hb\right)$
is $\text{sigmoid}\left(a_{i}+\sum_{k}W_{ik}h_{k}\right)$; and for
Gaussian data, the conditional density $P\left(x_{i}\mid\hb\right)$
is $\mathcal{N}\left(a_{i}+\sum_{k}W_{ik}h_{k};\mathbf{1}\right)$. 

Since $\hb$ is discrete, Mv.RBM can be considered as a mixture model
of $2^{K}$ components that shared the same parameter. This suggests
that Mv.RBM can be used for outlier detection in the same way that
GMM does. 

\subsection{Outlier Detection on Mixed-Type Data \label{subsec:Free-Energy}}

Recall that for outlier detection as in Eq.~(\ref{eq:outlier-decision})
we need the marginal distribution $P(\xb)=\sum_{\hb}P(\xb,\hb)$,
which is:
\[
P(\xb)\propto\sum_{\hb}\exp\left(-E(\xb,\hb)\right)=\exp\left(-F(\xb)\right)
\]
where $F(\xb)=-\log\sum_{\hb}\exp\left(-E(\xb,\hb)\right)$ is known
as \emph{free-energy}. Notice that the free-energy equals the negative
log-density up to an additive constant:
\[
F(\xb)=-\log P(\xb)+\text{constant}
\]
Thus \emph{we can use the free-energy as the outlier score} to rank
data instances, following the detection rule in Eq.~(\ref{eq:outlier-decision}).

\subsubsection{Computing free-energy}

Although estimating the free-energy amounts to summing over $2^{K}$
configurations of the hidden layer, we can still compute the summation
efficiently, thanks to the decomposition of the energy function in
Eq.~(\ref{eq:Mv.RBM-energy}). We can rewrite the free-energy as
follows:
\begin{align}
F(\xb) & =\sum_{i}E_{i}(x_{i})-\sum_{k}\log\left(1+\exp\left(b_{k}-\sum_{i}G_{ik}(x_{i})\right)\right)\label{eq:free-energy}
\end{align}
This free-energy can be computed in linear time.

\subsubsection{Controlling model expressiveness}

A major challenge of unsupervised outlier detection is the phenomenon
of \emph{swamping effect}, where an inlier is misclassified as outlier,
possibly due a large number of true outliers in the data \cite{serfling2014general}.
When data models are highly expressive \textendash{} such as large
RBMs and Mv.RBMs \textendash{} outliers are included by the models
as if they have patterns themselves, even if these patterns are weak
and differ significantly from the regularities of the inliers. One
way to control the model expressiveness is to limit the number of
hidden layers $K$ (hence the number of mixing components $2^{K}$).
Another way is to apply early stopping \textendash{} learning stops
before convergence has occurred.

\subsubsection{Summary}

To sum up, Mv.RBM, coupled with free-energy, offers a disciplined
approach to mixed-type outlier detection that meet three desirable
criteria: (i) c\emph{apturing correlation structure between types},
(ii) \emph{measuring deviation from the norm}, and (iii) \emph{efficient
to compute}.

\section{Experiments}

We present experiments on synthetic and real-world data. For comparison,
we implement well-known single-type outlier detection methods including
Gaussian mixture model (GMM), Probabilistic Principal Component Analysis
(PPCA) \cite{tipping1999probabilistic} and one-class SVM (OCSVM)
\cite{manevitz2001one}. The number of components of PPCA model is
set so that the discarded energy is the same as the anomaly rate in
training data. For OCSVM, we use radial basis kernel with $\nu=0.7$.
GMM and PPCA are probabilistic, and thus data log-likelihood can be
computed for outlier detection. 

Since all of these single-type methods assume numerical data, we code
nominal types using dummy binaries. For example, a \emph{A} in the
nominal set \{\emph{A,B,C}\} is coded as (1,0,0) and \emph{B} as (0,1,0).
This coding causes some nominal information loss, since the coding
does not ensure that only one value is allowed in nominal variables.
For all methods, the detection threshold is based on the $\alpha$
percentile of the training outlier scores. Whenever possible, we also
include results from other recent mixed-type papers, ODMAD \cite{koufakou2008detecting},
Beta mixture model (BMM) \cite{bouguessa2015practical} and GLM-t
\cite{lu2016discovering}. We followed the same mechanism they used
to generate outliers. 

\subsection{Synthetic Data}

\begin{figure}
\begin{centering}
\includegraphics[width=0.7\textwidth]{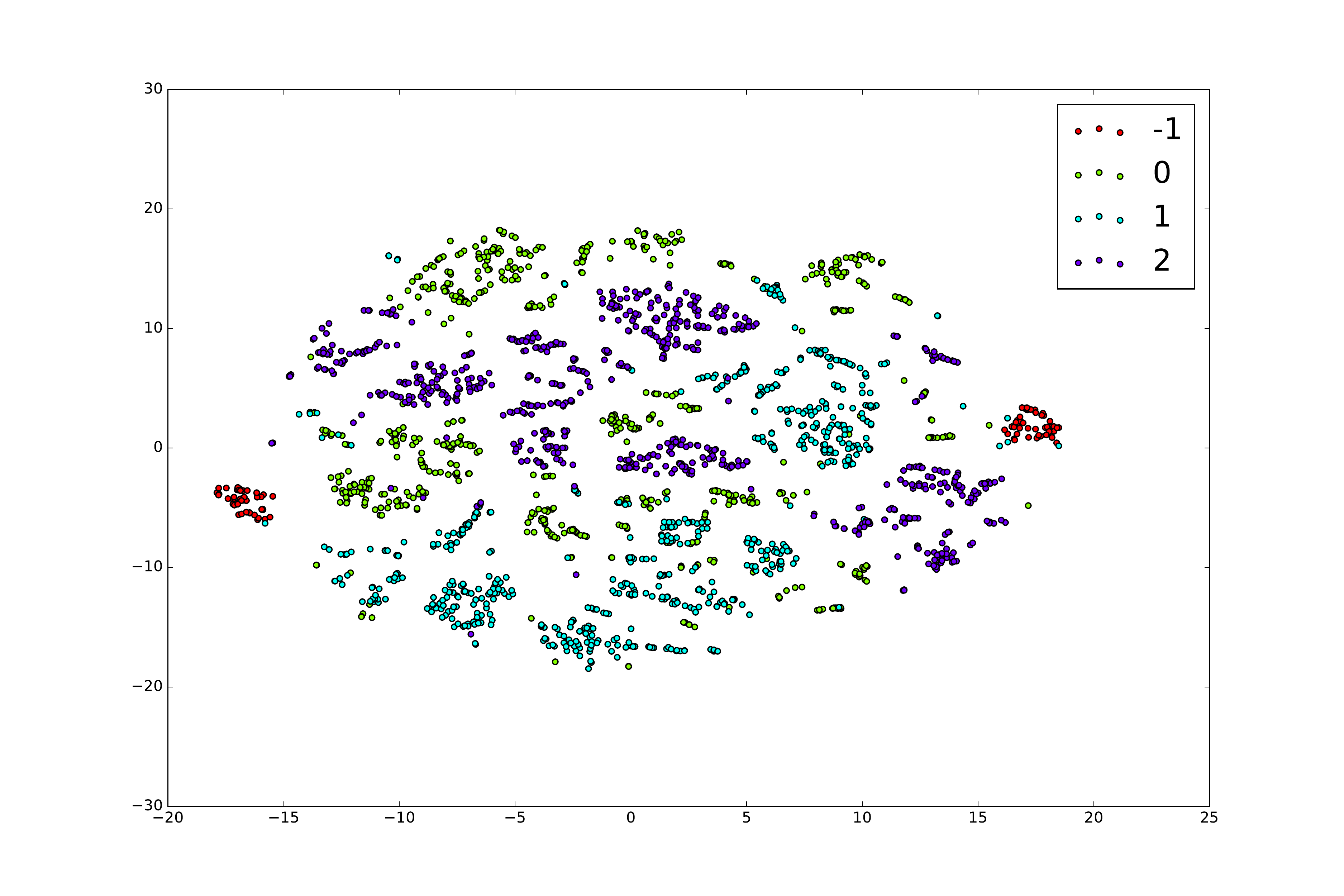}
\par\end{centering}
\caption{Synthetic data with 3 normal clusters (cluster IDs 0,1,2) and 1 set
of scattered outliers (ID: -1, colored in red). Best viewed in color.\label{fig:synthetic-data}}
\end{figure}

We first evaluate the behaviors of Mv.RBM on synthetic data with controllable
complexity. We simulate mixed-type data using a generalized Thurstonian
theory, where Gaussians serve as underlying latent variables for observed
discrete values. Readers are referred to \cite{truyen_phung_venkatesh_icml13}
for a complete account of the theory. For this study, the underlying
data is generated from a GMM of 3 mixing components with equal mixing
probability. Each component is a multivariate Gaussian distributions
of 15 dimensions with random mean and positive-definite covariance
matrix. From each distribution, we simulate 1,000 samples, creating
a data set size 3,000. To generate outliers, we randomly pick 5\%
of data, and add uniform noise to each dimension, i.e., $x_{i}\leftarrow x_{i}+e_{i}$
where $e_{i}\sim\mathcal{U}$. For visualization, we use t-SNE to
reduce the dimensionality to 2 and plot the data in Fig.~\ref{fig:synthetic-data}. 

Out of 15 variables, 3 are kept as Gaussian and the rest are used
to create mixed-type variables. More specifically, 3 variables are
transformed into binaries using random thresholds, i.e., $\tilde{x}_{i}=\delta(x_{i}\ge\theta_{i})$.
The other 9 variables are used to generate 3 nominal variables of
size 3 using the rule: $\tilde{x}_{i}=\arg\max\left(x_{i1},x_{i2},x_{i3}\right)$. 

\begin{figure}
\begin{centering}
\includegraphics[width=0.9\textwidth,height=0.5\textwidth]{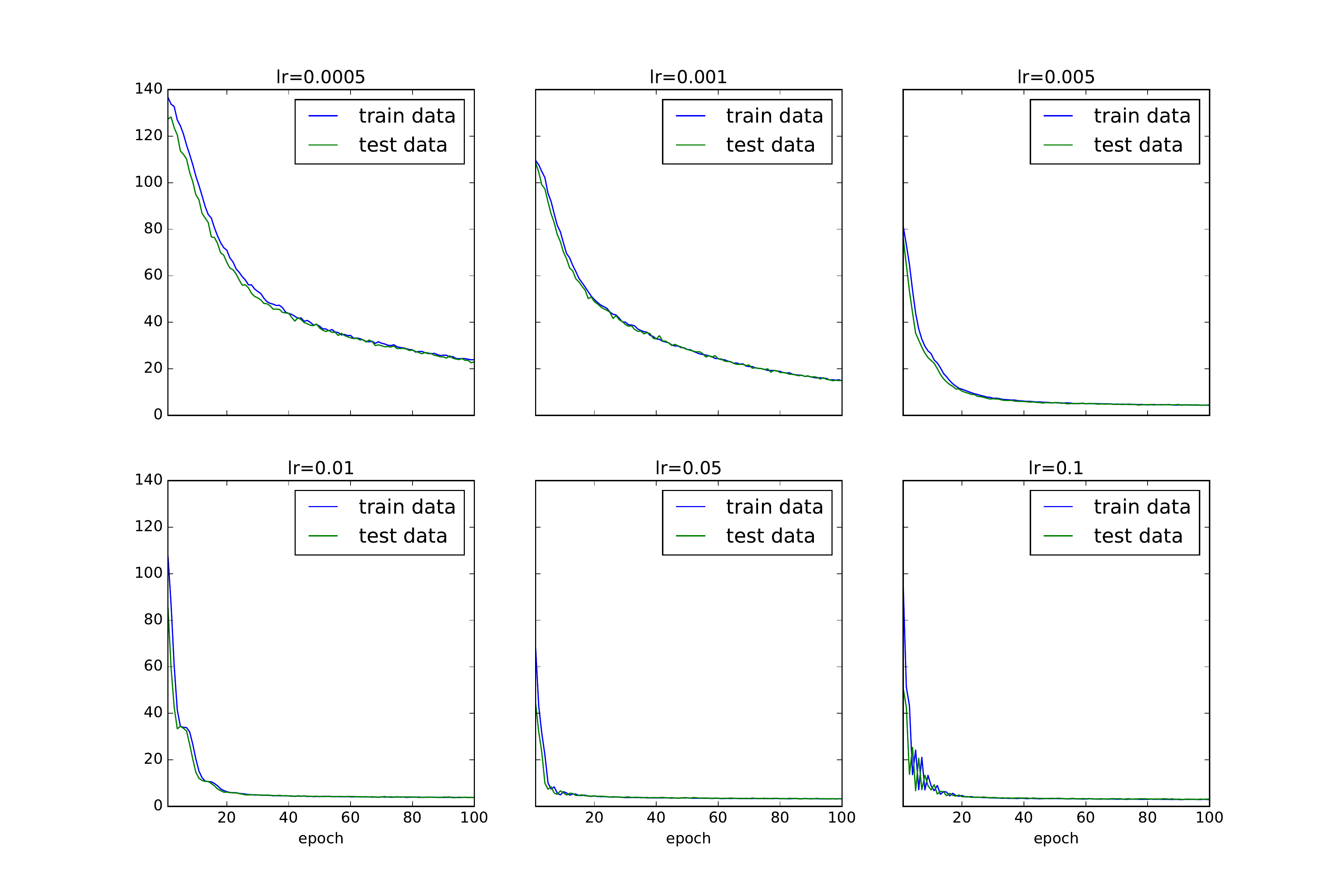}
\par\end{centering}
\caption{Learning curves of Mv.RBM (50 hidden units) on synthetic data for
different learning rates. The training and test curves almost overlap,
suggesting no overfitting. Best viewed in color.\label{fig:synthetic-learn}}
\end{figure}

Models are trained on 70\% data and tested on the remaining 30\%.
This testing scheme is to validate the generalizability of models
on unseen data. The learning curves of Mv.RBM are plotted in Fig.~\ref{fig:synthetic-learn}.
With the learning rate of 0.05, learning converges after 10 epochs.
No overfitting occurs.

\begin{figure}
\begin{centering}
\includegraphics[width=0.7\textwidth]{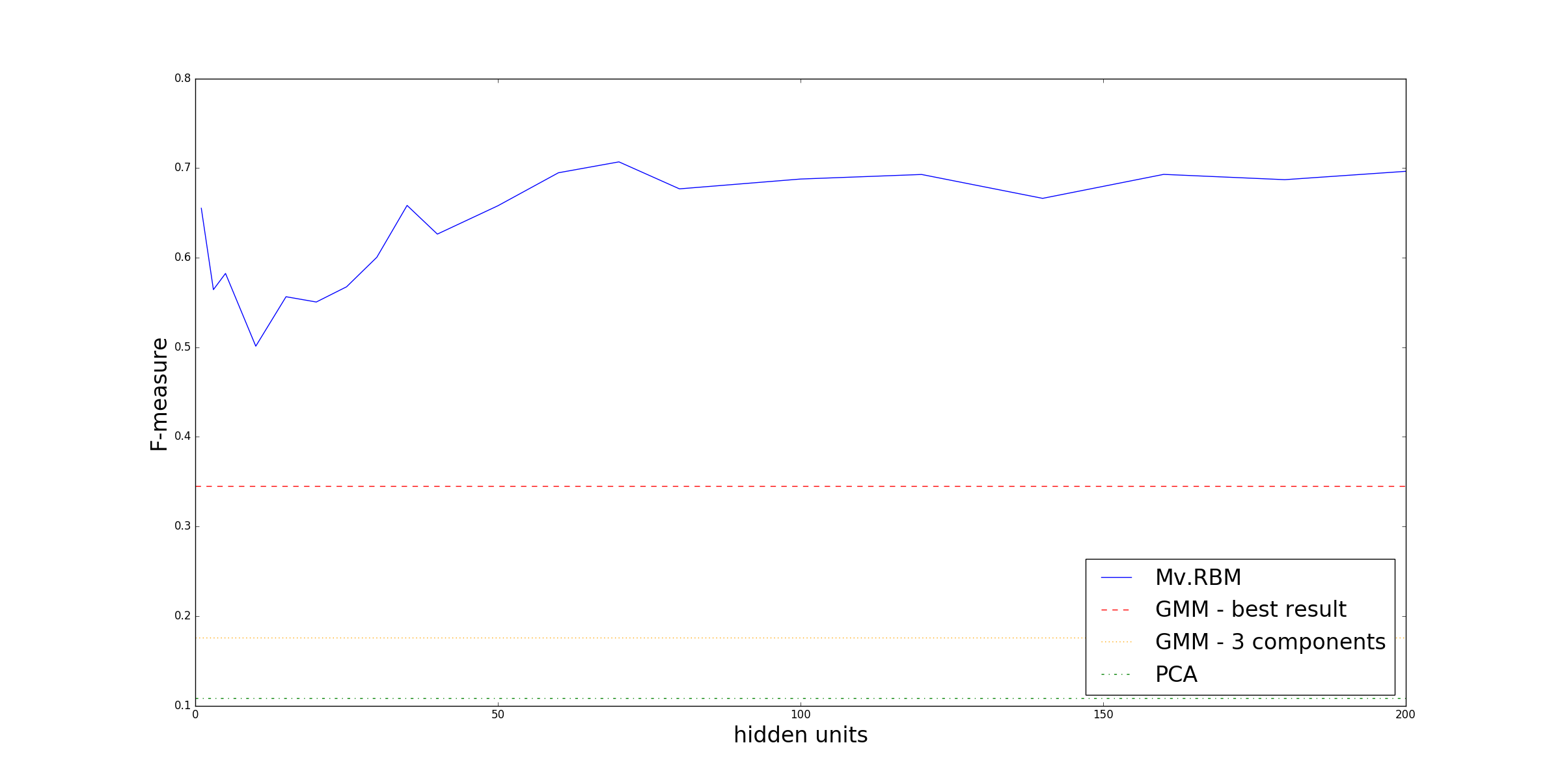}
\par\end{centering}
\caption{Performance of Mv.RBM in F-score on synthetic data as a function of
number of hidden units. Horizontal lines are performance measures
of PCA (in green) and GMM (best result in red; and with 3 components
in yellow). Best viewed in color.\label{fig:synthetic-perf}}
\end{figure}

The decision threshold $\beta$ in Eq.~(\ref{eq:outlier-decision})
is set at 5 percentile of the training set. Fig.~\ref{fig:synthetic-perf}
plots the outlier detection performance of Mv.RBM (in F-score) on
test data as a function of model size (number of hidden units). To
account for random initialization, we run Mv.RBM 10 times and average
the F-scores. It is apparent that the performance of Mv.RBM is competitive
against that of GMM. The best F-score achieved by GMM is only about
0.35, lower than the worst F-score by Mv.RBM, which is 0.50. The PCA
performs poorly, with F-score of 0.11, possibly because the outliers
does not conform to the notion of residual subspace assumed by PCA.

The performance difference between Mv.RBM and GMM is significant considering
the fact that the underlying data distribution is drawn from a GMM.
It suggests that when the correlation between mixed attributes is
complex like this case, using GMM even with the same number of mixture
components cannot learn well. Meanwhile, Mv.RBM can handle the mixed-type
properly, \emph{without knowing the underlying data assumption}. Importantly,
varying the number of hidden units does not affect the result much,
suggesting the stability of the model and it can free users from carefully
crafting this hyper-parameter.

\subsection{Real Data}

For real-world applications, we use a wide range of mixed-type datasets.
From the UCI repository\footnote{https://archive.ics.uci.edu/ml/datasets.html},
we select 7 datasets which were previously used as benchmarks for
mixed-type anomaly detection \cite{bouguessa2015practical,ghoting2004loaded,lu2016discovering}.
Data statistics are reported in Table~\ref{tab:dataset-prop}. We
generate outliers by either using rare classes whenever possible,
or by randomly injecting a small proportion of anomalies, as follows:
\begin{itemize}
\item \textbf{Using rare classes}: For the KDD99 \textit{10 percent} dataset
(KDD99-10), intrusions (outliers) account for 70\% of all data, and
thus it is not possible to use full data because outliers will be
treated as normal in unsupervised learning. Thus, we consider all
normal instances from the original data as inliers, which accounts
for 90\% of the new data. The remaining 10\% outliers are randomly
selected from the original intrusions. 
\item \textbf{Outliers injection}: For the other datasets, we treat data
points as normal objects and generate outliers based on a contamination
procedure described in \cite{bouguessa2015practical,lud2016iscovering}.
Outliers are created by randomly selecting 10\% of instances and modifying
their default values. For numerical attributes (Gaussian, Poisson),
values are shifted by 2.0 to 3.0 times standard deviation. For discrete
attributes (binary, categorical), the values are switched to alternatives.
\end{itemize}
Numerical attributes are standardized to zero means and unit variance.
For evaluation, we randomly select 30\% data for testing, and 70\%
data for training. Note that since learning is unsupervised, outliers
must also be detected in the training set since there are no ground-truths.
The outliers in the test set is to test the generalizability of the
models to unseen data. 

\begin{table}
\begin{centering}
\begin{tabular}{|l|r|r|c|c|c|c|c|}
\hline 
\multirow{2}{*}{Dataset} & \multicolumn{2}{c|}{No. Instances} & \multicolumn{5}{c|}{No. Attributes}\tabularnewline
\cline{2-8} 
 & \multicolumn{1}{c|}{Train} & Test & Bin. & Gauss. & Nominal & Poisson & Total\tabularnewline
\hline 
\hline 
\emph{KDD99-10} & 75,669 & 32,417 & 4 & 15 & 3 & 19 & 41\tabularnewline
\hline 
\emph{Australian Credit} & 533 & 266 & 3 & 6 & 5 & 0 & 14\tabularnewline
\hline 
\emph{German Credit} & 770 & 330 & 2 & 7 & 11 & 0 & 20\tabularnewline
\hline 
\emph{Heart} & 208 & 89 & 3 & 6 & 4 & 0 & 13\tabularnewline
\hline 
\emph{Thoracic Surgery} & 362 & 155 & 10 & 3 & 3 & 0 & 16\tabularnewline
\hline 
\emph{Auto MPG} & 303 & 128 & 0 & 5 & 3 & 0 & 8\tabularnewline
\hline 
\emph{Contraceptive} & 1136 & 484 & 3 & 0 & 4 & 1 & 8\tabularnewline
\hline 
\end{tabular}
\par\end{centering}
\caption{\label{tab:dataset-prop}Characteristics of mixed-type datasets. The
proportion of outliers are 10\%.}
\end{table}

\subsubsection{Models setup}

The number of hidden units in Mv.RBM is set to $K=2$ for the KDD99-10
dataset, and to $K=5$ for other datasets. The parameters of Mv. RBM
are updated using stochastic gradient descent, that is, update occurs
after every mini-batch of data points. For small datasets, the mini-batch
size is equal to the size of the entire datasets while for KDD99-10,
the mini-batch size is set to 500. The learning rate is set to 0.01
for all small datasets, and to 0.001 for KDD99-10. Small datasets
are trained using momentum of 0.8. For KDD99-10, we use Adam \cite{kingma2014adam},
with $\beta_{1}=0.85$ and $\beta_{2}=0.995$. For small datasets,
the number of mixture components in GMM is chosen using grid search
in the range from 1 to 30 with a step size of 5. For KDD99-10, the
number of mixture components is set to 4.

\subsubsection{Results}

\begin{figure}
\begin{centering}
\begin{tabular}{cc}
\includegraphics[width=0.6\textwidth,height=0.4\textwidth]{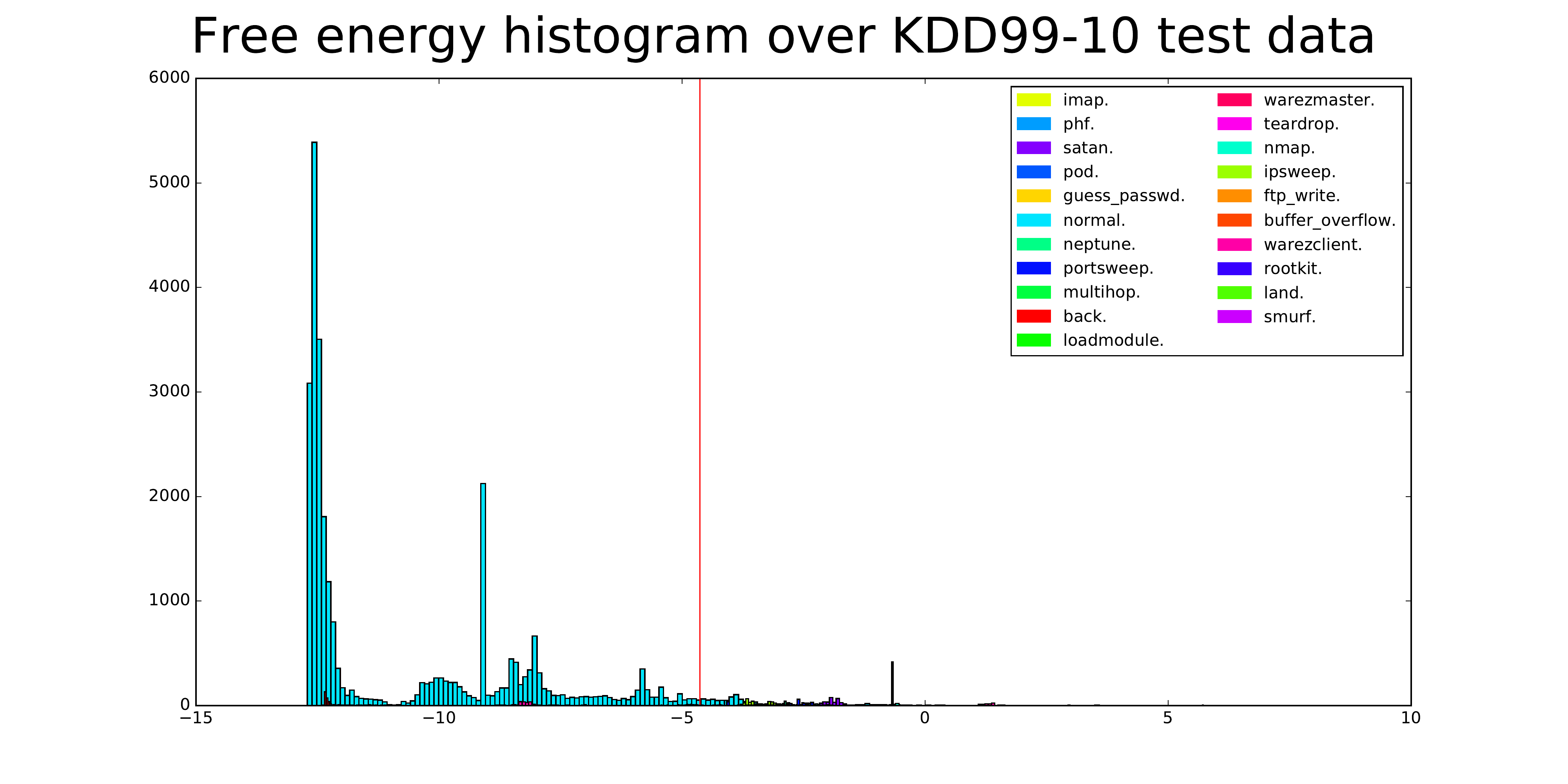} & \includegraphics[width=0.4\textwidth,height=0.4\textwidth]{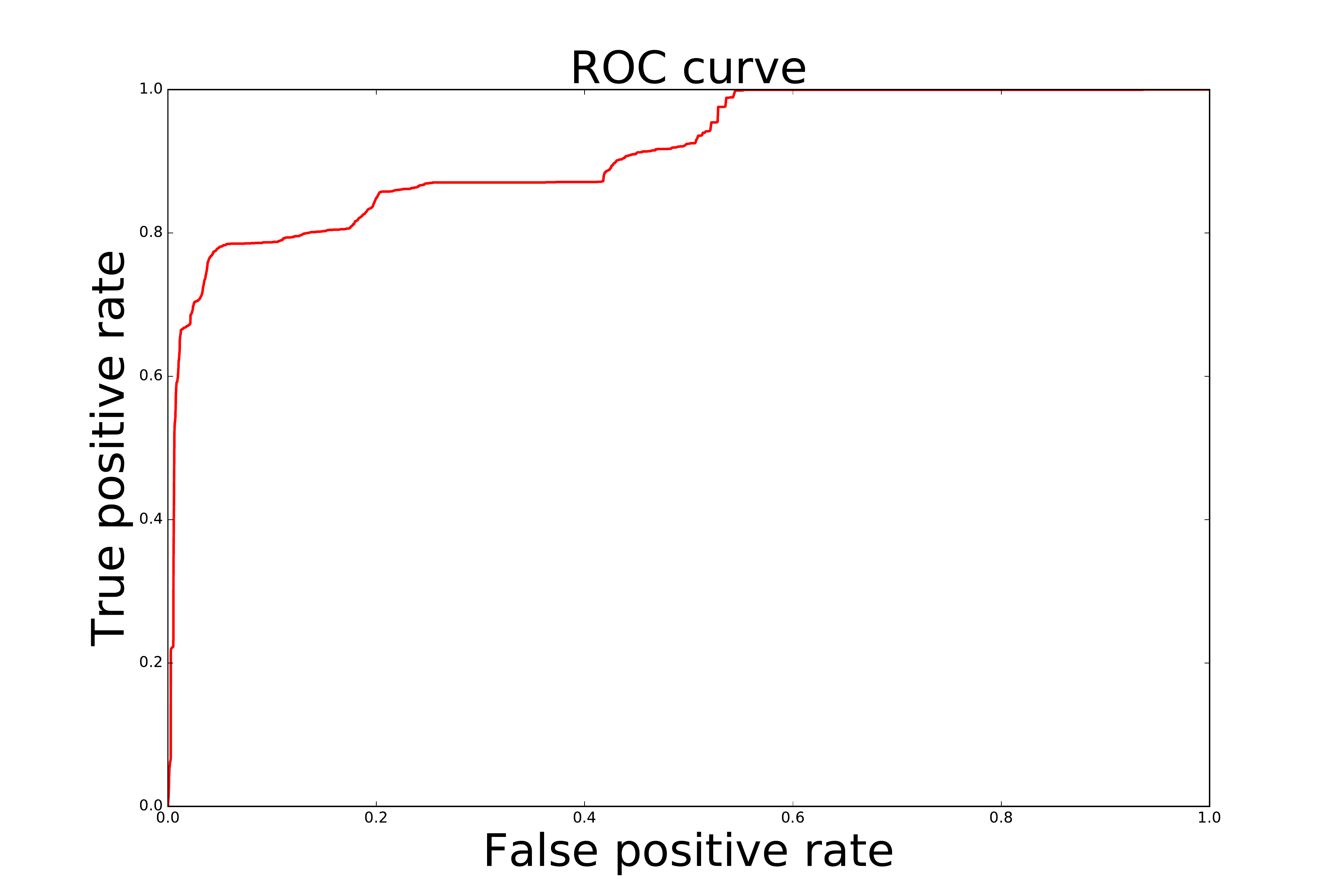}\tabularnewline
(a) Free-energy histogram & (b) ROC curve\tabularnewline
\end{tabular}
\par\end{centering}
\caption{Outlier detection on the KDD99-10 dataset. (a) Histogram of free-energies.
The vertical line separates data classified as inliers (left) from
those classified as outliers (right). The color of majority (light
blue) is inlier. Best viewed in color. (b) ROC curve (AUC = 0.914).
\label{fig:Histogram-of-free-energy}}
\end{figure}

Fig.~\ref{fig:Histogram-of-free-energy}(a) shows a histogram of
free-energies computed using Eq.~(\ref{eq:free-energy}) on the KDD99-10
dataset. The inliers/outliers are well-separated into the low/high
energy regions, respectively. This is also reflected in an Area Under
the ROC Curve (AUC) of 0.914 (see Fig.~\ref{fig:Histogram-of-free-energy}(b)).

\begin{table}
\begin{centering}
\begin{tabular}{|l|ccc|ccc|c|}
\hline 
\multirow{2}{*}{Dataset} & \multicolumn{3}{c|}{Single type} & \multicolumn{4}{c|}{mixed-type}\tabularnewline
\cline{2-8} 
 & GMM & OCSVM & PPCA & BMM & ODMAD & GLM-t & Mv.RBM\tabularnewline
\hline 
\hline 
\emph{KDD99-10} & 0.42 & 0.54 & 0.55 & \textendash{} & \textendash{} & \textendash{} & \textbf{0.71}\tabularnewline
\hline 
\emph{Australian Credit} & 0.74 & 0.84 & 0.38 & \textbf{0.972} & 0.942 & \textendash{} & 0.90\tabularnewline
\hline 
\emph{German Credit} & 0.86 & 0.86 & 0.02 & 0.934 & 0.810 & \textendash{} & \textbf{0.95}\tabularnewline
\hline 
\emph{Heart} & 0.89 & 0.76 & 0.64 & 0.872 & 0.630 & 0.72 & \textbf{0.94}\tabularnewline
\hline 
\emph{Thoracic Surgery} & 0.71 & 0.71 & 0.70 & \textbf{0.939} & 0.879 & \textendash{} & 0.90\tabularnewline
\hline 
\emph{Auto MPG} & \textbf{1.00} & \textbf{1.00} & 0.67 & 0.625 & 0.575 & 0.64 & \textbf{1.00}\tabularnewline
\hline 
\emph{Contraceptive} & 0.62 & 0.84 & 0.02 & 0.673 & 0.523 & \textendash{} & \textbf{0.91}\tabularnewline
\hline 
\hline 
\textbf{\emph{Average}} & \emph{0.75} & \emph{0.79} & \emph{0.43} & \emph{0.84} & \emph{0.73} & \emph{0.68} & \textbf{\emph{0.91}}\tabularnewline
\hline 
\end{tabular} 
\par\end{centering}
\caption{Outlier detection F-score. \label{tab:F-measure}}
\end{table}

The detection performance in term of F-score on test data is reported
in Tables~\ref{tab:F-measure}. The mean of all single type scores
is 0.66, of all competing mixed-type scores is 0.77, and of Mv.RBM
scores is 0.91. These demonstrate that (a) a proper handling of mixed-types
is required, and (b) Mv.RBM is highly competitive against other mixed-type
methods for outlier detection. Point (a) can also be strengthened
by looking deeper: On average, the best competing mixed-type method
(BMM) is better than the best single-type method (OCSVM). For point
(b), the gaps between Mv.RBM and other methods are significant: On
average, Mv.RBM is better than the best competing method \textendash{}
the BMM (mixed-type) \textendash{} by 8.3\%, and better than the worst
method \textendash{} the PPCA (single type), by 111.6\%. On the largest
dataset \textendash{} the KDD99-10 \textendash{} Mv.RBM exhibits a
significant improvement of 29.1\% over the best single type method
(PPCA).

\section{Discussion}

This paper has introduced a new method for mixed-type outlier detection
based on an energy-based model known as Mixed-variate Restricted Boltzmann
Machine (Mv.RBM). Mv.RBM avoids direct modeling of correlation between
types by using binary latent variables, and in effect, model the correlation
between each type and the binary type. We derive free-energy, which
equals the negative log of density up-to a constant, and use it as
the outlier score. Overall, the method is highly scalable. Our experiments
on mixed-type datasets of various types and characteristics demonstrate
that the proposed method is competitive against the well-known baselines
designed for single types, and recent models designed for mixed-types.
These results (a) support the hypothesis that in mixed-data, proper
modeling of types should be in place for outlier detection, and (b)
show Mv.RBM is a powerful density-based outlier detection method.

Mv.RBM opens several future directions. First, Mv.RBM transforms multiple
types into a single type through its hidden posteriors. Existing single-type
outlier detectors can be readily employed. Second, Mv.RBM can serve
as a building block for deep architecture, such as Deep Belief Networks
and Deep Boltzmann Machine. It would be interesting to see how deep
networks perform in non-prediction settings such as outlier detection.

\section*{Acknowledgments}

This work is partially supported by the Telstra-Deakin Centre of Excellence
in Big Data and Machine Learning.

\bibliographystyle{plain}

\end{document}